\documentclass[11pt,a4paper]{article}
\usepackage[hyperref]{acl2021}
\usepackage{times}
\usepackage{latexsym}
\usepackage{graphicx}
\usepackage{algorithmic}
\usepackage[ruled]{algorithm2e}
\usepackage{amsmath}
\usepackage{booktabs}
\DeclareMathOperator*{\argmax}{arg\,max}

\usepackage{amsfonts}
\usepackage{multirow}

\usepackage{microtype}
\usepackage{tikz}
\newcommand*\circled[1]{\tikz[baseline=(char.base)]{
            \node[shape=circle,draw,inner sep=2pt] (char) {#1};}}

\aclfinalcopy % Uncomment this line for the final submission
\def\aclpaperid{107} %  Enter the acl Paper ID here

\title{SocAoG: Incremental Graph Parsing for Social Relation Inference in Dialogues}

\author{Liang Qiu{\normalfont\textsuperscript{1}}, Yuan Liang{\normalfont\textsuperscript{2}}, Yizhou Zhao{\normalfont\textsuperscript{1}}, Pan Lu{\normalfont\textsuperscript{1}}, Baolin Peng{\normalfont\textsuperscript{3}}, \\
{\bf Zhou Yu{\normalfont\textsuperscript{4}}, Ying Nian Wu{\normalfont\textsuperscript{1}}, Song-Chun Zhu{\normalfont\textsuperscript{1}}} \\
\textsuperscript{1}UCLA Center for Vision, Cognition, Learning, and Autonomy \\
\textsuperscript{2}University of California, Los Angeles \\
\textsuperscript{3}Microsoft Research, Redmond \\
\textsuperscript{4}University of California, Davis \\
\texttt{liangqiu@ucla.edu}
}
\date{}

\begin{document}
\maketitle
\begin{abstract}
Inferring social relations from dialogues is vital for building emotionally intelligent robots to interpret human language better and act accordingly. We model the social network as an And-or Graph, named SocAoG, for the consistency of relations among a group and leveraging attributes as inference cues. Moreover, we formulate a sequential structure prediction task, and propose an $\alpha$--$\beta$--$\gamma$ strategy to incrementally parse SocAoG for the dynamic inference upon any incoming utterance:  \textit{(i)} an $\alpha$ process predicting attributes and relations conditioned on the semantics of dialogues, \textit{(ii)} a $\beta$ process updating the social relations based on related attributes, and \textit{(iii)} a $\gamma$ process updating individual's attributes based on interpersonal social relations. Empirical results on DialogRE and MovieGraph show that our model infers social relations more accurately than the state-of-the-art methods. Moreover, the ablation study shows the three processes complement each other, and the case study demonstrates the dynamic relational inference.
% \footnote{The code is released at \url{https://github.com/Liang-Qiu/SocAoG-dialogues}.}
\end{abstract}

\section{Introduction}
Social relations form the basic structure of our society, defining not only our self-images but also our relationships~\citep{sztompka2002socjologia}. Robots with a higher emotional quotient (EQ) have the potential to understand users' social relations better and act appropriately. Given a dialogue as context and a set of entities, the task of Dialogue Relation Extraction (DRE) predicts the relation types between the entities from a predefined relation set. Table \ref{tab:dialogre} shows such an example from the dataset DialogRE~\citep{yu-etal-2020-dialogue}. 
\begin{table}[]
\centering
\resizebox{0.9\linewidth}{!}{%
\begin{tabular}{llll}
\toprule
\textbf{S1:} & \multicolumn{3}{l}{\begin{tabular}[c]{@{}l@{}}Well then we’ll-we’ll see you the day after tomorrow. \\ Mom?! Dad?! What-what…what you guys doing here?!\end{tabular}} \\
\midrule
\textbf{S2:} & \multicolumn{3}{l}{\begin{tabular}[c]{@{}l@{}}Well you kids talk about this place so much, we thought \\ we’d see what all the fuss is about.\end{tabular}}           \\
\midrule
\textbf{S3:} & \multicolumn{3}{l}{I certainly see what the girls like coming here.}                                                       \\
\midrule
\textbf{S1:} & \multicolumn{3}{l}{Why?!}                                                                                                  \\
\midrule
\textbf{S3:} & \multicolumn{3}{l}{The sexy blonde behind the counter.}                                                                    \\
\midrule
\textbf{S1:} & \multicolumn{3}{l}{Gunther?!}                                                                                              \\
\midrule
\textbf{S2:} & \multicolumn{3}{l}{Your mother just added him to her list.}                                                                \\
\midrule
\textbf{S1:} & \multicolumn{3}{l}{What? Your-your list?}                                                                                  \\
\midrule
\textbf{}    & \textbf{Argument Pair}                    & \textbf{Trigger}                   & \textbf{Relation Type}                    \\
\textbf{R1}  & (S2, S1)                                  & dad                                & per:children                              \\
\textbf{R2}  & (S3, Gunther)                             & sexy blonde                        & per:positive\_impression                   \\
\textbf{R3}  & (S3, S1)                                  & mom                                & per:children                              \\
\textbf{R4}  & (S1, S3)                                  & mom                                & per:parents                               \\
\textbf{R5}  & (S1, S2)                                  & dad                                & per:parents                               \\
\bottomrule
\end{tabular}%
}
\caption{A dialogue example from DialogRE~\citep{yu-etal-2020-dialogue}. Trigger word annotations are not used for training, but rather for illustrating purpose only.}
\label{tab:dialogre} 
\vspace{-5mm}
\end{table}

\begin{figure*}[]
\begin{center}
\centerline{\includegraphics[width=0.9\linewidth]{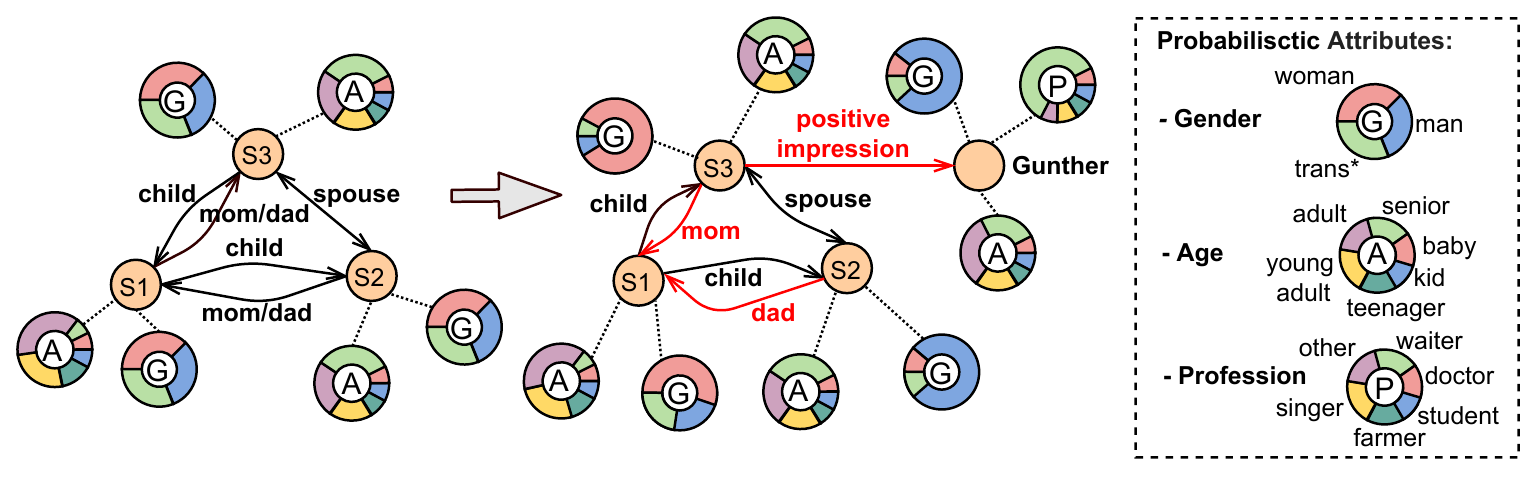}}
\caption{Our method iteratively updates the robot's belief of users' individual attributes and social relations, similar to human's reasoning process. The left and right graph show the established and updated belief, respectively. }
\label{fig:social_relation}
\end{center}
\vspace{-10mm}
\end{figure*}

Existing researches using BERT-based models~\citep{devlin2018bert, yu-etal-2020-dialogue, xue2020embarrassingly} or graph-based models~\citep{xue2020gdpnet, chen2020dialogue} focus on identifying entities' relations from the semantics of dialogues---they utilize either the attention mechanism or a refined token graph to locate informative words (\textit{e.g.}, ``dad" and ``mom") that imply the argument pairs' relations. However, there are still three missing parts in current models for social relation inference according to our observations. First, current models lack the explicit modeling of the relational consistency among a group of people---such consistency helps humans reason about the social relation of two targets by using their relations with a third person. For the example in Table \ref{tab:dialogre}, by knowing S2 and S3 are S1's parents and S3 is S1's mother, we can infer that S2 is S1's dad. Second, the personal attribute cues (\textit{e.g.}, gender and profession) can also aid the relational inference but are not fully utilized. In the above example, besides inferring S3 is S1' mother according to S3's feminine attribute, we can also have a guess that Gunther is a waiter, which might be useful for the future social-relational inference. Third, since the BERT-based and token-graph-based models take dialogues as a whole for relation prediction, they cannot perform dynamic inference---updating the relational belief with an incoming dialogic utterance. This can limit their ability to track the evolving relations along social interactions, \textit{e.g.}, strangers become friends over a good chat~\citep{kukleva2020learning}, unveiling intermediate reasoning results, or dealing with long dialogues. 

Motivated by these observations, we propose to model social relation as an attributed And-Or graph (AoG)~\citep{zhu1998filters, zhu2007stochastic, wu2011numerical, shu2016learning, qi2018human}, named SocAoG, and develop an incremental graph parsing algorithm to jointly infer human attributes and social relations from a dialogue. In specific, SocAoG describes social relations and personal attributes with contextual constraints of groups and hierarchical representations. To incrementally parse SocAoG and track social relations, we apply Markov Chain Monte Carlo (MCMC) to sample from the posterior probability calculated by three complementary processes ($\alpha$--$\beta$--$\gamma$)~\citep{qu2020few, zayaraz2015concept}. Figure \ref{fig:social_relation} schematically demonstrates a graph update of both relations (\textit{i.e.}, disambiguating mom/dad and adding a new party) and attributes (\textit{e.g.}, gender and profession) with the utterance ``\textbf{S2}: \textit{Your mother just added him to the list.}" from the example dialogue in Table \ref{tab:dialogre}. 

We evaluate our method on two datasets of DialogRE~\citep{yu-etal-2020-dialogue} and MovieGraph~\cite{vicol2018moviegraphs} for relation inference, and the results show that our method outperforms the state-of-the-art (SoTA) ones. Overall, we make the following contributions: (\textit{i}) We propose to model and infer social relations and individual's attributes jointly with SocAoG for the consistency of attributes and social relations among a group. To the best of our knowledge, it is the first time done in the dialogue domain; (\textit{ii}) The MCMC sampling from $\alpha$--$\beta$--$\gamma$ posterior enables dynamic inference---incrementally parsing the social relation graph, which can be useful for tracking relational evolution, reflecting the reasoning process, and handling long dialogues; (\textit{iii}) We perform an ablation study on each process of $\alpha$--$\beta$--$\gamma$ to investigate the information contribution, and perform case studies to show the effectiveness of our dynamic reasoning. 

\section{Related Work}
We review the related works on the social relation inference from documents, which is a well-studied task, and those from dialogues, which is the emerging task that our work is focused on. 

\subsection{Relation Inference from Documents}
Most of the existing literature focus on relation extraction from professional edited news reports or websites. They typically output a set of ``subject-predicate-object" triples after reading the entire document~\citep{bach2007review, mintz-etal-2009-distant, kumar2017survey}. While early works mostly utilize feature-based methods~\citep{kambhatla-2004-combining, miwa-sasaki-2014-modeling, gormley-etal-2015-improved} and kernel-based methods~\citep{zelenko2003kernel, zhao-grishman-2005-extracting, mooney2006subsequence},
more recent studies use deep learning methods such as recurrent neural networks or transformers~\citep{kumar2017survey}. For example, \citet{zhou-etal-2016-attention-based} propose bidirectional LSTM model to capture the long-term dependency between entity pairs, \citet{zhang-etal-2017-position} present PA-LSTM to encode global position information, and \citet{alt-etal-2019-fine, papanikolaou-etal-2019-deep} fine-tune pre-trained transformer language models for relation extraction.

Two streams of work are closely related to our method. Regarding social network modeling, while most works treat pairs of entities isolated \cite{yu-etal-2020-dialogue, xue2020gdpnet, chen2020dialogue}, \citet{srivastava2016inferring} formulate the interpersonal relation inference as structured prediction~\citep{belanger2016structured, qiu-etal-2020-structured, zhao2020vertical}, inferring the collective assignment of relations among all entities from a document~\citep{li-etal-2020-high, jin2020relation}. Regarding relation evolution, a few works are aimed to learn the dynamics in social networks, \textit{i.e.,} the development of relations, from narratives by Hidden Markov Models~\citep{chaturvedi2017unsupervised}, Recurrent Neural Networks~\citep{kim-klinger-2019-frowning}, deep recurrent autoencoders~\citep{iyyer-etal-2016-feuding}. Our method differs from the aforementioned works by modeling the structured social relations and their changes concurrently, which can be useful for the task of tracking social network evolution \cite{doreian1997dynamics} and unveiling the reasoning process of relations. We achieve this by parsing the graph incrementally per utterance with the proposed $\alpha$--$\beta$--$\gamma$ strategy. 

\begin{figure*}[ht]
\begin{center}
\centerline{\includegraphics[width=0.9\linewidth]{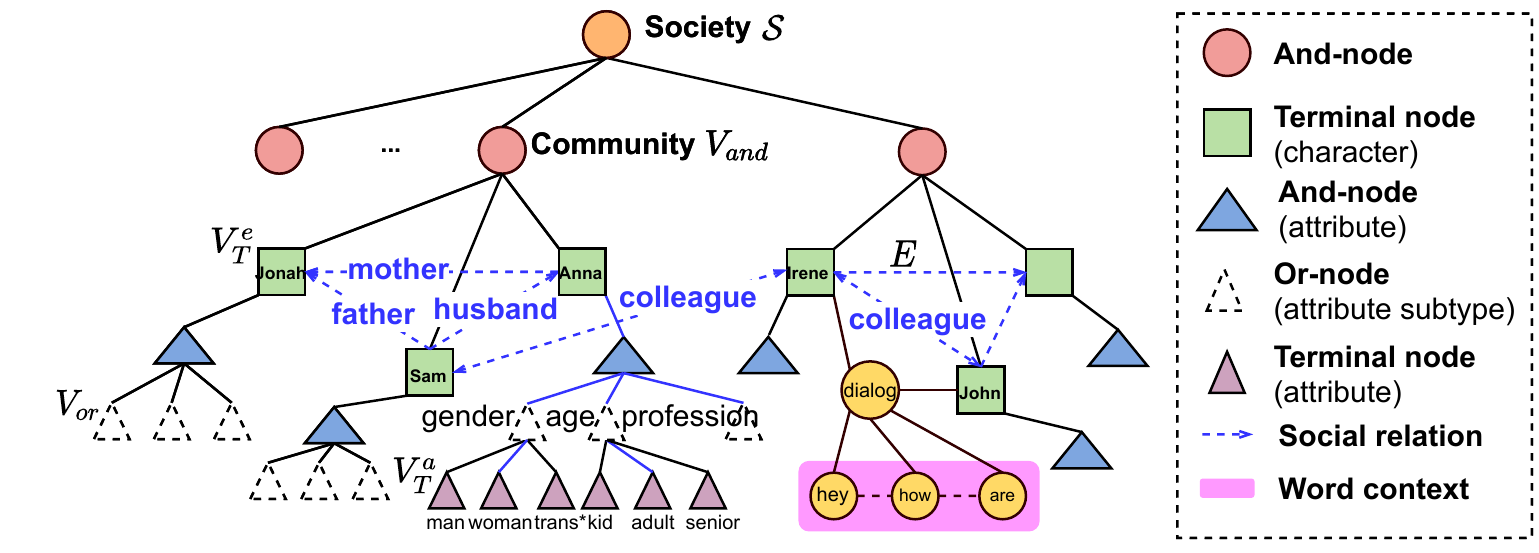}}
\caption{\textbf{SocAoG}: Attributed And-Or Graph representation of a social network. A parse graph determining each attribute and relation type is marked in blue lines. Dialogues are governed by the word context and associated human attributes and relations.}
\label{fig:aog}
\end{center}
\vspace{-10mm}
\end{figure*}
\subsection{Relation Inference from Dialogues}
Recently,~\citet{yu-etal-2020-dialogue} introduce the first human-annotated dialogue-based relation extraction dataset DialogRE, in which relations are annotated between arguments that appear in a dialogue session. Compared with traditional relation extraction tasks, DialogRE emphasizes the importance of tracking speaker-related information within the context across multiple sentences. SoTA methods can be categorized into token-graph models and pre-trained language models. For typical token-graph models, \citet{chen2020dialogue} present a token graph attention network, and \citet{xue2020gdpnet} further generate a latent multi-view graph to capture relationships among tokens, which is then refined to select important words for relation extraction. For pre-trained models, \citet{yu-etal-2020-dialogue} evaluate a BERT-based baseline model~\citep{devlin2018bert} and a modified version BERTs, which takes speaker arguments into consideration. \citet{xue2020embarrassingly} propose a simple yet effective BERT-based model, SimpleRE, that takes a novel input format to capture the interrelations among all pairs of entities. 

Both categories of SoTA models take a discriminative approach, whereas ignoring two key constraints on relations: \textit{(i)} social relation consistency in a group and \textit{(ii)} human attributes. 
Different from them, our method formulates the task as dialogue generation from an attributed relation graph, so that the posterior relation estimation models both two constraints. 
Moreover, SoTA models also assume the relations are static---they cannot learn the dynamics of the relations, while the incremental graph updating strategy naturally enables the dynamic relation inference. 
% Compared with the traditional relation extraction tasks, DialogRE emphasizes the importance of speaker-related information in dialog-based cross-sentence, which can make it more challenging. 
% Compared with the traditional relation extraction from documents, DialogRE requires the reasoning of attitude and context from the utterances among multiple entities across a dialog, which makes it more challenging. 
% Early works of relation extraction include feature-based methods~\citep{kambhatla-2004-combining, miwa-sasaki-2014-modeling, gormley-etal-2015-improved} and kernel-based methods~\citep{zelenko2003kernel, zhao-grishman-2005-extracting, mooney2006subsequence}. More recent works on learning-based models improve the RE performance to another level. For example, \citet{zhou-etal-2016-attention-based} propose bidirectional LSTM model to capture the long-term dependency between entity pairs. \citet{zhang-etal-2017-position} present PA-LSTM to encode global position information to boost the performance. 

\section{Problem Formulation}
Our goal is to construct a social network through utterances in dialogue. The network is a heterogeneous physical system~\citep{yongqiang1997theory} with particles representing entities and different types of edges representing social relations. Each entity is associated with multiple types of attributes, while each type of relation is governed by a potential function defined in human attribute and value space, acting as the social norm. The relations are often asymmetric, \textit{e.g.}, A is B's father does not mean B is A's father. To model the network, we utilize an attributed And-Or Graph (A-AoG), a probabilistic grammar model with attributes on nodes. Such design takes advantage of the reconfigurability of its probabilistic context-free grammar to reflect the alternative attributes and relations, and the contextual relations defined on Markov Random Field to model the social norm constraints.

The social network graph, named SocAoG, is diagrammatically shown in Figure \ref{fig:aog}. Formally, SocAoG is defined as a 5-tuple:
\begin{equation}
    \mathcal{G}=<S,V,E,X,P>
\end{equation}
, where $S$ is the root node for representing the interested society. $V=V_{and} \cup V_{or} \cup V_T^e \cup V_T^a$ denotes all nodes' collection. Among them, And-nodes $V_{and}$ represent the set of social communities, which can be decomposed to a set of entity terminal nodes, $V_T^e$, representing human members. Community detection is based on the social network analysis \cite{bedi2016community, du2007community}, and can benefit the modeling of loosely connected social relations. Each human entity is associated with an And-node that breakdowns the attributes into subtypes such as gender, age, and profession.
All the subtypes consist of an Or-node set, $V_{or}$, for representing branches to alternatives of attribute values. Meanwhile, all the attribute values are represented as a set of terminal nodes $V_T^a$. 
We denote $E$ to be the edge set describing social relations, $X(v_i)$ to be the attributes associated with node $v_i$, and $X(\vec{e}_{ij})$ to be the social relation type of edge $\vec{e}_{ij} \in E$. 

Given $P$ to be the probability model defined on SocAoG, a parse graph $pg$ is an instantiation of SocAoG with determined attribute selections for every Or-node and relation types for every edge. For a dialogue session with $T$ turns $D_T=\{D^{(1)},D^{(2)},...,D^{(T)}\}$, where $D^{(t)}$ is the utterance at turn $t$, our method infers the attributes and social relations incrementally over turns:
\begin{equation}
    \mathcal{G}_T=\{pg^{(1)},pg^{(2)},...,pg^{(T)}\}
\end{equation}
, where $pg^{(t)}$ represents the belief of SocAoG at the dialogue turn $t$. We incrementally update the $pg$ by maximizing the posterior probability:
\begin{equation}
\label{eqn:posterior}
    pg^*=\argmax_{pg}p(pg|D;\theta)
\end{equation}
, where $pg^*$ is the optimum social relation belief, and $\theta$ is the set of model parameters. 

\section{Algorithm}
\subsection{$\alpha$--$\beta$--$\gamma$ for Graph Inference}
For simplicity, we denote $X(v_i)$ as $\mathbf{v}_i$ and $X(\vec{e}_{ij})$ as $\mathbf{e}_{ij}$ in the rest of the paper. We introduce three processes, \textit{i.e.}, $\alpha$, $\beta$, and $\gamma$ process, to infer any SocAoG belief $pg^*$. We start by rewriting the posterior probability as a Gibbs distribution:  
\begin{equation}
\label{eqn:bayes}
    \begin{aligned}
    p(pg|D;\theta) &\propto p(D|pg;\theta)p(pg;\theta) \\
                   &=\frac{1}{Z} exp \{-\mathcal{E}(D|pg;\theta) -\mathcal{E}(pg;\theta)\}
    \end{aligned}
\end{equation}
, where $Z$ is the partition function. $\mathcal{E}(D|pg;\theta)$ and $\mathcal{E}(pg;\theta)$ are dialogue- and social norm-based energy potentials respectively, measuring the cost of assigning a graph instantiation.

Denoting a dialogue as a sequence of words: $D=\{w_1,w_2,...,w_\mathcal{T}\}$, the dialogue likelihood energy term $\mathcal{E}(D|pg;\theta)$ can be expressed with a language model conditioned on the parse graph:
\begin{equation}
\label{eqn:alpha}
    \begin{aligned}
        \mathcal{E}(D|pg;\theta) =& \sum_{t=1}^{\mathcal{T}}\mathcal{E}(w_t|\mathbf{c}_t, pg) \\
                                 =& \sum_{t=1}^{\mathcal{T}}-\log(p(w_t|\mathbf{c}_t, pg))
    \end{aligned}
\end{equation}
, where $\mathbf{c}_t=[w_1,...,w_{t-1}]$ is the context vector. Intuitively, the word selection depends on the word context, the entities' attributes and their interpersonal relations. 
% As such, we learn the conditional causal language model by finetuning a (TODO: BERT!)GPT-2~\citep{radford2019language} with a customized input format $\langle v_{i_0} \textbf{e}_{{i_0}{j_0}} v_{j_0} ... v_{i_n} \textbf{e}_{{i_n}{j_n}} v_{j_n} v_0 \textbf{v}_0 ... v_n \textbf{v}_n D\rangle$, which is a concatenation of flattened parse graph string encoding the current belief and dialogue history. 
We approximate the likelihood by finetuning a BERT-based transformer with a customized input format $\langle \texttt{[CLS]} \texttt{D} \texttt{[SEP]} v_{i_0} \textbf{e}_{{i_0}{j_0}} v_{j_0} ... v_{i_n} \textbf{e}_{{i_n}{j_n}} v_{j_n} v_0 \textbf{v}_0 ... v_n$ $\textbf{v}_n \texttt{[SEP]}\rangle$, which is a concatenation of the dialogue history $\texttt{D}$ and a flattened parse graph string encoding the current belief. We call the estimation of $pg$ from the dialogue likelihood $p(w_t|\mathbf{c}_t, pg)$ to be the \textbf{$\alpha$ process}. $\alpha$ process lacks the explicit constraints for social norms related to interpersonal relations and human attributes. 

\begin{figure}[ht]
    \begin{center}
    \centerline{\includegraphics[width=0.8\linewidth]{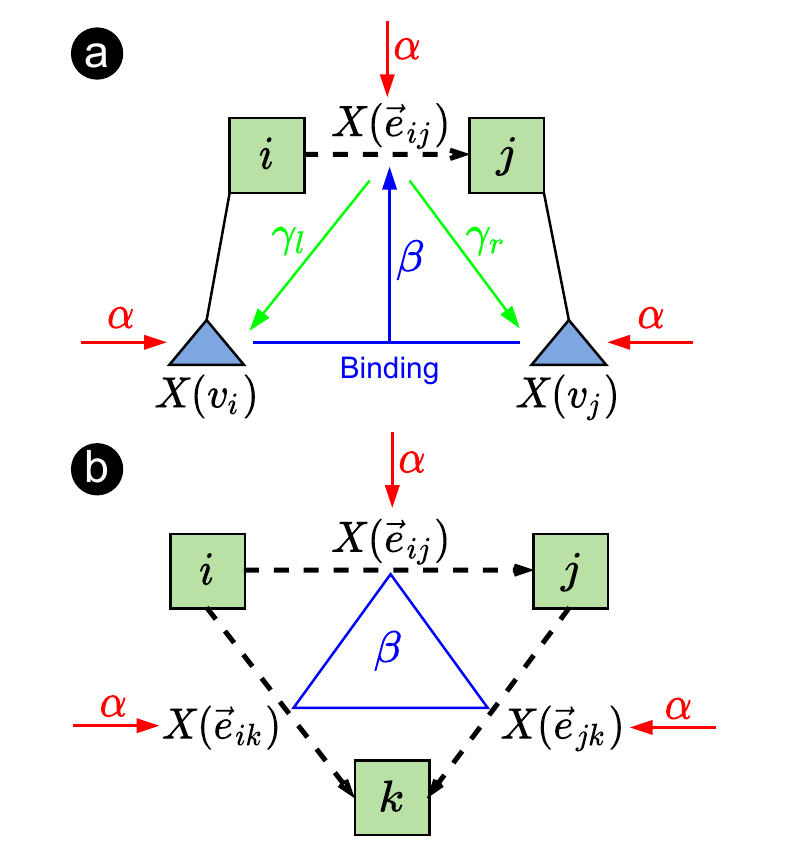}}
    \caption{(a) $\alpha$--$\beta$--$\gamma$ process for SocAoG. (b) $\alpha$--$\beta$ process for reduced SocAoG without attributes. Note that this $\beta$ is only modeling the interrelations among $X(\vec{e})$.}
    \label{fig:3process}
    \end{center}
    \vspace{-8mm}
\end{figure}

For the social norm-based potential, we design it to be composed of three potential terms:
\begin{equation}
\label{eqn:beta-gamma}
    \begin{aligned}
    \mathcal{E}(pg;\theta) = &-\beta\sum_{v_i,v_j\in V(pg)} \log(p(\mathbf{e}_{ij}|\mathbf{v}_{i},\mathbf{v}_{j})) \\
                             &-\gamma_l\sum_{\vec{e}_{ij}\in E(pg)} \log(p(\mathbf{v}_{i}|\mathbf{e}_{ij})) \\
                             &-\gamma_r\sum_{\vec{e}_{ij}\in E(pg)} \log(p(\mathbf{v}_{j}|\mathbf{e}_{ij}))
    \end{aligned}
\end{equation}
, where $V(pg)$ and $E(pg)$ are the set of terminal nodes and relations in the parse graph, respectively.
We call the term $p(\mathbf{e}_{ij}|\mathbf{v}_{i},\mathbf{v}_{j})$ the \textbf{$\beta$ process}, in which we bind the attributes of node $v_i$ and $v_j$ to update their relation edge $\mathbf{e}_{ij}$, in order to model the constraint on relations from human attributes. 
Reversely, we call the terms $p(\mathbf{v}_{i}|\mathbf{e}_{ij})$ and $p(\mathbf{v}_{j}|\mathbf{e}_{ij})$ the \textbf{$\gamma$ process}, in which we use the social relation edge $\mathbf{e}_{ij}$ to update the attributes of node $v_i$ and $v_j$. This models the impact of relation to the attributes of related entities. 
$\beta, \gamma_l$, and $\gamma_r$ are weight factors balancing $\alpha$, $\beta$ and $\gamma$ processes. Figure \ref{fig:3process}(a) shows the graph inference schema with the three processes. Combining equation \ref{eqn:bayes}, \ref{eqn:alpha}, and \ref{eqn:beta-gamma}, we get a posterior probability estimation $ p(pg|D;\theta)$ of parse graph $pg$, with the guarantee of the attribute and social norm consistencies. 

Here we also provide a reduced version of our model, SocAoG\textsubscript{reduced}, which applies when characters' attributes annotation are not available for training\footnote{Both SocAoG and SocAoG\textsubscript{reduced} do not need attribute annotation during inference once trained.}. With the same dialogue-based energy potential, We define the parse graph prior energy over a set of relation triangles:
\begin{equation}
    \mathcal{E}(pg;\theta) = -\beta\sum_{\vec{e}_{ij},\vec{e}_{ik},\vec{e}_{jk}\in E(pg)} \log(p(\mathbf{e}_{ij}|\mathbf{e}_{ik},\mathbf{e}_{jk})). 
\end{equation}
The method directly models the constraint of two entities' relation from their relations to others, with the inference schema demonstrated in Figure \ref{fig:3process}(b). 

\subsection{Incremental Graph Parsing}
Incrementally parsing the SocAoG is accomplished by repeatedly sampling a new parse graph $pg^{(t)}$ from the posterior probability $p(pg^{(t)}|D^{(t)};\theta)$. We utilize a Markov Chain Monte Carlo (MCMC) sampler to update our parse graph since the complexity of the problem caused by multiple energy terms. 
\begin{algorithm}[t]
\caption{Incremental SocAoG Parsing for Social Relation Inference}
\label{alg:parsing}
\small
\begin{algorithmic}
\STATE {\bfseries Input:} dialogue $D_T=\{D^{(1)},D^{(2)},...,D^{(T)}\}$, \\
                        \hskip1.0em target argument pairs $\{a_1,a_2\}$.
\STATE {\bfseries Initialize} $pg^{(0)}$. Initialize $\mathbf{v}_i$ and $\mathbf{e}_{ij}$. \\
\FOR{$t=1,...,T$}
\FOR{$s=1,...,S$}
    \STATE Compute the posterior $p(pg|D^{(t)};\theta)$.
    \STATE Make proposal moves with probabilities \\
    \hskip1.0em $q_1,q_2$ to get a new parse graph $pg'$.
    \STATE Compute the posterior $p(pg'|D^{(t)};\theta)$.
    \STATE Compute acceptance rate \\
    \hskip1.0em $\alpha(pg'|pg, D^{(t)};\theta)$.
    \STATE Accept/reject $pg'$ according to the \\
    \hskip1.0em acceptance rate.
\ENDFOR
\STATE \Return $\mathbf{e}_{a_1,a_2}$ from the average of accepted \\
\hskip1.0em $pg$ samples.
\ENDFOR
\end{algorithmic}
\end{algorithm}

% We start from a random SocAoG parse graph $pg^{(0)}$ as the initialization. 
% At each dialogue turn $t$, the last $pg$ we sampled in dialogue turn $t-1$ will be used to initialize the parse graph so that they form a continuous Markov chain. 
At each dialogue turn $t$, we initialize the parse graph with the $\alpha$ classification process, by replacing all the Or-Node tokens with a special token $\texttt{[CLS]}$. We sample the parse graph for $S$ steps and use the average value of obtained samples as an approximation of $pg^{(t)}$. We design two types of Markov chain dynamics used at random probabilities $q_i, i=1,2$ to make proposal moves:
\begin{itemize}
\item Dynamics $q_1$: randomly pick a relation edge $\vec{e}_{ij}$ under the uniform distribution, flip its social relation type $\mathbf{e}_{ij}$ according to the prior distribution given by $\beta$ process:
\begin{equation}
    \prod_{v_i,v_j\in V(pg)} p(\mathbf{e}_{ij}|\mathbf{v}_i,\mathbf{v}_j).
\end{equation} 
\item Dynamics $q_2$: randomly pick a terminal node $v_i$ and its attribute subtype under the uniform distribution, and flip the one-hot value of attribute $\mathbf{v}_i$ according to the prior distribution given by $\gamma$ process: 
\begin{equation}
    \prod_{\vec{e}_{ij}\in E(pg)} p(\mathbf{v}_i|\mathbf{e}_{ij}) \prod_{\vec{e}_{ji}\in E(pg)} p(\mathbf{v}_i|\mathbf{e}_{ji}).
\end{equation} 
\end{itemize}
Using the Metropolis-Hastings algorithm~\citep{chib1995understanding}, the proposed new parse graph $pg'$ is accepted according to the following acceptance probability:
\begin{equation}
    \begin{aligned}
    \alpha(pg'|pg, D;\theta) &=\min(1, \frac{p(pg'|D;\theta)p(pg|pg')}{p(pg|D;\theta)p(pg'|pg)}) \\
                             &=\min(1, \frac{p(pg'|D;\theta)}{p(pg|D;\theta)})
    \end{aligned}
    \label{formula_acc}
\end{equation}
, where the proposal probability rate is cancelled out since the proposal moves are symmetric in probability. We summarize the incremental SocAoG parsing in Algorithm \ref{alg:parsing}. 
Dialogues give a continuously evolving energy landscape: at the beginning of iterations, $p(pg^{(0)}|D;\theta)$ is a ``hot" distribution with a large energy value; by iterating the $\alpha$--$\beta$--$\gamma$ processes for $pg$ updates through the dialogue, the $pg$ converges to the $pg^*$, which is much cooler. 
% However, the personal attributes and social relations are constrained by multiple energy terms; hence they are too complicated to sample directly. 
% Here, we utilize a Markov Chain Monte Carlo (MCMC) sampler to update our parse graph from the posterior defined by the SocAoG. 
% However, the personal attributes and social relations are constrained by multiple energy terms; hence they are too complicated to sample directly. 
% The last $pg$ we sampled in dialogue turn $t-1$ will be used to initialize the parse graph in dialogue turn $t$ so they can form a continuous Markov chain. 

\section{Experiments}
\label{sec:exp}
\begin{table*}[]
\small
\centering
\resizebox{0.9\textwidth}{!}{%
\begin{tabular}{l|cc|cc|c|c}
\toprule
               & \multicolumn{4}{c|}{\textbf{DialogRE} (\textbf{V2})}                   & \multicolumn{2}{c}{\textbf{MovieGraph}} \\
               & \multicolumn{2}{c|}{\textbf{Dev}} & \multicolumn{2}{c|}{\textbf{Test}} & \textbf{Dev}            & \textbf{Test}           \\
\textbf{Methods}        & $\mathbf{F}1 (\sigma)$        & $\mathbf{F}1_c (\sigma)$      & $\mathbf{F}1 (\sigma)$        & $\mathbf{F}1_c (\sigma)$       & $\mathbf{F}1 (\sigma)$           & $\mathbf{F}1 (\sigma)$           \\
\midrule
BERT~\citep{devlin2018bert}           & 59.4 (0.7)  & 54.7 (0.8) & 57.9 (1.0)  & 53.1 (0.7)  & 50.6 (1.2)  & 53.6 (0.3)              \\
BERT\textsubscript{S}~\citep{yu-etal-2020-dialogue} & 62.2 (1.3)  & 57.0 (1.0) & 59.5 (2.1)  & 54.2 (1.4)  & 50.7 (1.1) & 53.6 (0.4)    \\
GDPNet~\citep{xue2020gdpnet}         & 67.1 (1.0)  & 61.5 (0.8) & 64.3 (1.1)  & 60.1 (0.9)  & 53.1 (1.1) & 56.4 (0.8)              \\
SimpleRE~\citep{xue2020embarrassingly}       & 68.2 (1.1)  & 63.4 (0.6) & 66.7 (0.7) & 63.3 (0.9) & 55.2 (0.5)  & 58.1 (0.7)      \\
SocAoG\textsubscript{reduced} (our method) & 69.1 (0.4) & 65.7 (0.5) & 68.6 (0.9) & 65.4 (1.1) & \textbf{60.7 (0.4)}  & 63.2 (0.3)              \\
SocAoG (our method)          &\textbf{69.5 (0.8)}  &\textbf{66.1 (0.7)} & \textbf{69.1 (0.5)} & \textbf{66.5 (0.8)} & 60.1 (0.6) & \textbf{64.1 (0.8)}              \\
\bottomrule
\end{tabular}%
}
\caption{Performance comparison between BERT, BERT\textsubscript{S}, GDPNet, SimpleRE, SocAoG\textsubscript{reduced}, and SocAoG. We report 5-run average results and the standard deviation ($\sigma$).}
\label{tab:result}
\vspace{-2mm}
\end{table*}

\subsection{Datasets}
We use DialogRE (V2)\footnote{https://github.com/nlpdata/dialogre}~\citep{yu-etal-2020-dialogue} and MovieGraph\footnote{http://moviegraphs.cs.toronto.edu/}~\citep{vicol2018moviegraphs} for evaluating our method. Detailed descriptions on the two datasets, \textit{e.g.}, relation and attribute types, are provided in Appendix \ref{sec:appendix}. 

% contains 10,168 relational triples from the American comedy series \textit{Friends}.
DialogRE contains 36 relation types (17 of them are interpersonal) that exist between pairs of arguments. For the joint parsing of relation and attribute, we further annotate the entity arguments with attributes from four subtypes (by following the practice of MovieGraph~\citep{vicol2018moviegraphs}): gender, age, profession, and ethnicity, according to Friends Central in Fandom\footnote{https://friends.fandom.com/wiki/Friends\_Wiki}. DialogRE is split into training (1073), validation (358), and test (357). Following previous works~\citep{yu-etal-2020-dialogue,xue2020gdpnet}, we report macro $\mathbf{F}1$ scores in both the standard and conversational settings ($\mathbf{F}1_c$).

MovieGraph provides graph-based annotations of social situations from 51 movies. Each graph comprises nodes representing the characters, their emotional and physical attributes, relationships, and interactions. We use a subset (40) of MovieGraph with available full transcripts and split the dataset into training (26), validation (6), and test (8). For MovieGraph, we only evaluate with $\mathbf{F}1$ since the trigger word annotation for computing $\mathbf{F}1_c$ is not available.

\subsection{Experiment Settings}
We learn the SocAoG model with a contrastive loss~\citep{hadsell2006dimensionality} comparing the posterior of a positive parse graph against a negative one. All parameters are learned by gradient descent using the Adam optimizer~\citep{kingma2014adam}. During the inference stage, for each utterance, we run the MCMC 
% for 200 warm-up (burn-in) steps and then sample 
for $S=\min\{w\times(KM+K(K-1)N), S_{max}\}$ steps given $K$ entities, $M$ attributes, $N$ relations, and a sweep number of $w$. The probability of flipping the relation $q_1$ is set to 0.7 to bias towards the relation prediction at first. 
\subsection{Baseline Models}
We compare our method with both transformer-based (\textbf{BERT}, \textbf{BERT\textsubscript{S}}, \textbf{SimpleRE}) and graph-based (\textbf{GDPNet}) models. Given dialogue history $D$ and target argument pair $(v_i,v_j)$, \textbf{BERT}~\citep{devlin2018bert} takes input sequences formatted as $\langle\texttt{[CLS]} d \texttt{[SEP]} v_i \texttt{[SEP]} v_j \texttt{[SEP]}\rangle$. 
\textbf{BERT\textsubscript{S}}~\citep{yu-etal-2020-dialogue} is a speaker-aware modification of BERT, which also takes speaker information into consideration by converting it into a special token. 
\textbf{SimpleRE}~\citep{xue2020embarrassingly} models the relations between each pair of entities with a customized input format. \textbf{GDPNet}~\citep{xue2020gdpnet} takes in token representations from BERT and constructs a multi-view graph with a Gaussian Graph Generator. The graph is then refined through graph convolution and DTWPool to identify indicative words. 
% the interrelations among all relations in a text sequence. 
% Given dialogue history $D$, a set of subject entities $v_i=\{v_{i_0},v_{i_1},...,v_{i_n}\}$ and a set of object entities $v_j=\{v_{j_0},v_{j_1},...,v_{j_n}\}$, 
% the BERT input is formatted as $\langle\texttt{[CLS]} d \texttt{[SEP]} v_{i_0} \texttt{[CLS]} v_{j_0} \texttt{[SEP]},...,\texttt{[SEP]}$ $v_{i_n} \texttt{[CLS]} v_{j_n} \texttt{[SEP]}\rangle$. Except the first $\texttt{[CLS]}$, the subsequent $\texttt{[CLS]}$ tokens capture the relations between each pair of entities and are used for classification. 
% Given the same input, for the sentences that contain $v_i$ or $v_j$, the text indicating speaker \textit{e.g.,} ``Speaker 1", is replaced by a special token, \texttt{[S\textsubscript{1}]} or \texttt{[S\textsubscript{2}]}.
% Given a set of subject entities $v_i=\{v_{i_0},v_{i_1},...,v_{i_n}\}$ and a set of object entities $v_j=\{v_{j_0},v_{j_1},...,v_{j_n}\}$, 
% the model input is formatted as $\langle\texttt{[CLS]} d \texttt{[SEP]} v_{i_0} \texttt{[CLS]} v_{j_0} \texttt{[SEP]},...,\texttt{[SEP]}$ $v_{i_n} \texttt{[CLS]} v_{j_n} \texttt{[SEP]}\rangle$.
% $\texttt{[CLS]}$ tokens capture the relations between each pair of entities and are predicted with classification. 
% For fair comparison on the dataset without attribute annotation, we also report the performance of the reduced \textbf{SocAoG} model.
\begin{figure*}[ht]
\begin{center}
\centerline{\includegraphics[width=0.9\linewidth]{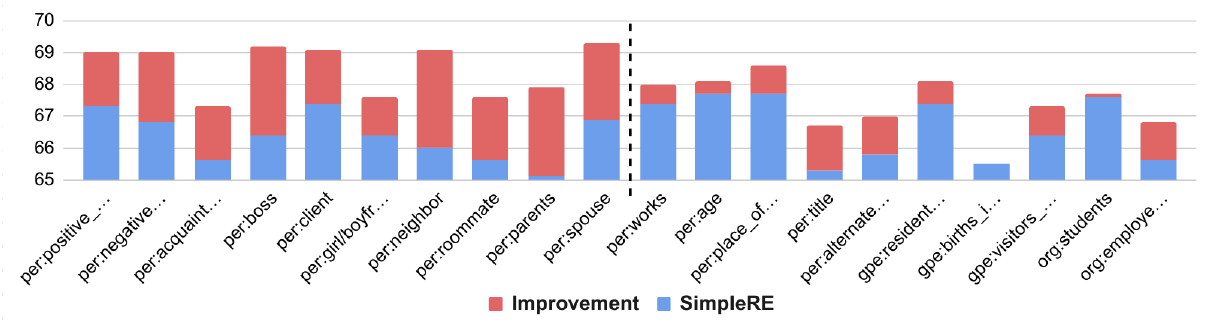}}
\caption{Performance boosts ($\mathbf{F}1$) of SocAoG compared to SimpleRE~\citep{xue2020embarrassingly} by relation type. The left bars to the dashed line are relations between humans, while the right ones are those between human and non-human entities.}
\label{fig:rel}
\end{center}
\vspace{-6mm}
\end{figure*}

\subsection{Performance Comparison}
\begin{figure*}[ht]
\begin{center}
\centerline{\includegraphics[width=\linewidth]{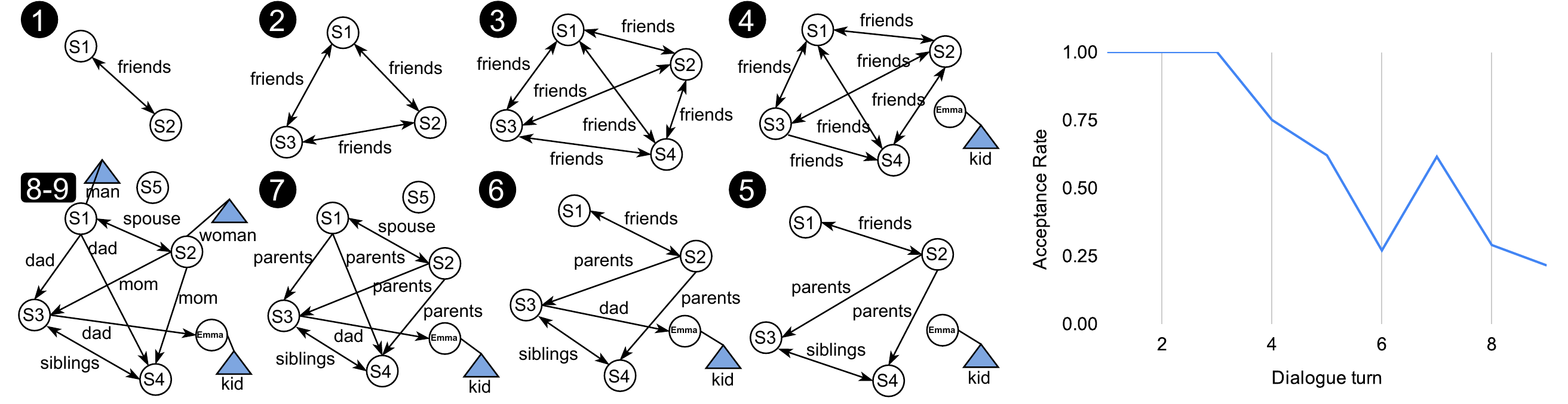}}
\caption{Left: inferred parse graph sequence from SocAoG based on the test dialogue in Table \ref{tab:case}. Note that dad/mom are not distinguished in DialogRE. Right: model convergence measured by acceptance rate at each dialogue turn.}
\label{fig:pg}
\end{center}
\vspace{-6mm}
\end{figure*}
Table \ref{tab:result} shows the performance comparison between different methods on the two datasets. 
It clearly shows that both of our models, SocAoG and SocAoG\textsubscript{reduced}, outperform the existing methods by all the metrics. In specific, without using any additional information of attributes, SocAoG\textsubscript{reduced} surpasses the state-of-the-art method (SimpleRE) by 1.9\% ($\mathbf{F}1$)/2.1\% ($\mathbf{F}1_c$) on DialogRE testing set, and by 5.1\% ($\mathbf{F}1_c$) on MovieGraph testing set. Such improvement shows the importance of relational consistency for the modeling, and proves the effectiveness of our SocAoG formulation to introduce the social norm constraints. 
% Supplementary materials include example cases where relational consistency is achieved by SocAoG\textsubscript{reduced} but failed for the existing methods. 

Moreover, by comparing between SocAoG and SocAoG\textsubscript{reduced}, we see that SocAoG further improves most of the metrics by leveraging the attribute information for relation reasoning, \textit{e.g.}, 69.1\% \textit{vs.} 68.6\% for DialogRE testing $\mathbf{F}1$ and 64.1\% \textit{vs.} 63.2\% for MovieGraph testing $\mathbf{F}1$. The results demonstrate our method can effectively take advantage of the attributes as cues for social relation predictions. 
We compare our SocAoG model with the existing model of highest accuracy (SimpleRE) by relation types, and see consistent improvements for all types. A part of the results are shown in Figure \ref{fig:rel}. 
We also observe that there are larger accuracy boosts for relations between human entities than non-human entities (\textit{e.g.}, human-place), by an average of $+$2.5\% \textit{vs.} $+$1.8\% in $\mathbf{F}1$, which is also reflected from Figure \ref{fig:rel} (left 10 bars \textit{vs.} right 10 bars). 
This can be explained as relation/attribute constraints are more meaningful for interpersonal relations, \textit{e.g.}, there are more constraints for the relation between three humans than the relation between two humans and a place.

Table \ref{tab:result} also sees more accuracy improvement on MovieGraph dataset than DialogRE ($+$3.2\% \textit{vs.} $+$6.0\% in test $\mathbf{F}1_c$ using SimpleRE as baseline). This is possibly because the dynamic inference nature of our method makes it effective for dealing with dialogues with more turns: while existing methods either truncate dialogues or use sliding windows, our method continuously updates the relation graph given an incoming turn. We case study the dynamic inference in the next subsection. 
% The complete comparison results are shown in Table \ref{tab:result}. We analyze the results on DialogRE and MovieGraph, respectively.
% \textbf{Results on DialogRE}. SocAoG surpasses the state-of-the art method by 1.3\%/2.8\% $\mathbf{F}1$ scores, and 2.4\%/2.4\% $\mathbf{F}1_c$ scores in both validation and test sets. GDPNet is superior to BERT and BERT\textsubscript{S} because of the graph refinement procedure for identifying informative words. SimpleRE learns the representations of all target relations and takes their inter-correlation into consideration. 
% Nonetheless, our algorithm outperforms them because it leverages important cues like individual attributes and it can process longer dialogue history that might be truncated by BERT-based models. Note that even without the additional attribute information, the reduced SocAoG algorithm still outperforms the baselines. We also compare the $\alpha$--$\beta$--$\gamma$ method with SimpleRE for each relation type and find our method is especially effective for interpersonal relation types, as shown in Figure \ref{fig:rel}.
% \textbf{Results on MovieGraph}. Since trigger word annotation is not available on MovieGraph, we only compute and report $\mathbf{F}1$ here. We use the proposed algorithm to process the whole transcript of each movie and predict the characters' relations. We can see the $\alpha$--$\beta$--$\gamma$ outperforms the baseline models by a large margin because of our algorithm's superiority in processing long sequences.

\subsection{Case Study on Dynamic Inference}
\begin{table}[]
\centering
\resizebox{0.9\linewidth}{!}{%
\begin{tabular}{llll}
\toprule
\circled{1}~~~~\textbf{S1, S2:} & \multicolumn{3}{l}{Hi!} \\
\midrule
\circled{2}~~~~\textbf{S3:} & \multicolumn{3}{l}{Hey!}           \\
\midrule
\circled{3}~~~~\textbf{S4:} & \multicolumn{3}{l}{So glad you came!}                   \\
\midrule
\circled{4}~~~~\textbf{S1:} & \multicolumn{3}{l}{I can't believe Emma is already one!}                         \\
\midrule
\circled{5}~~~~\textbf{S2:} & \multicolumn{3}{l}{\begin{tabular}[c]{@{}l@{}}I remember your first birthday! \\ Ross was jealous of all the attention we were giving you. \\ He pulled on his testicles so hard! \\ We had to take him to the emergency room!\end{tabular}} \\
\midrule
\circled{6}~~~~\textbf{S3:} & \multicolumn{3}{l}{There's something you didn't know about your dad!}            \\
\midrule
\circled{7}~~~~\textbf{S5:} & \multicolumn{3}{l}{Hey Mr. and Mrs. Geller! Let me help you with that.}          \\
\midrule
\circled{8}~~~~\textbf{S1:} & \multicolumn{3}{l}{Thank you!}   \\
\midrule
\circled{9}~~~~\textbf{S5:} & \multicolumn{3}{l}{\begin{tabular}[c]{@{}l@{}}Oh man, this is great, uh? The three of us together again! \\ You know what would be fun? \\ If we gave this present to Emma from all of us!\end{tabular}} \\
\bottomrule
\end{tabular}%
}
\caption{Dialogue example from the testing set of DialogRE~\citep{yu-etal-2020-dialogue}.}
\label{tab:case} 
\vspace{-6mm}
\end{table}
Our method incrementally updates the relation and attribute information for a group of entities upon per utterance input with the proposed $\alpha$--$\beta$--$\gamma$ strategy. 
Such dynamic inference can potentially help reflect the evolving relations, unveil the reasoning process, and deal with long dialogues.  
Figure \ref{fig:pg} shows the parse graph sequence by SocAoG inferring from a DialogRE testing dialogue as shown in Table \ref{tab:case}. 
We can see that the method continuously refines the relation/attributes from an initial guess with incoming contexts, \textit{e.g.} S2-S3: friends$\rightarrow$parents in turn 5.
Besides, the case also shows that attributes can aid relation predictions, \textit{e.g.}, the inferred age of Emma clarifies her relation with S3. 
Moreover, since our method models the relation consistency among a group, it can predict the relation between two humans that do not talk directly. 
For example, S1 and S2 are inferred to be a couple by their dialogues with S5 in turn 7. 

\begin{figure}[ht]
\begin{center}
\centerline{\includegraphics[width=0.9\linewidth]{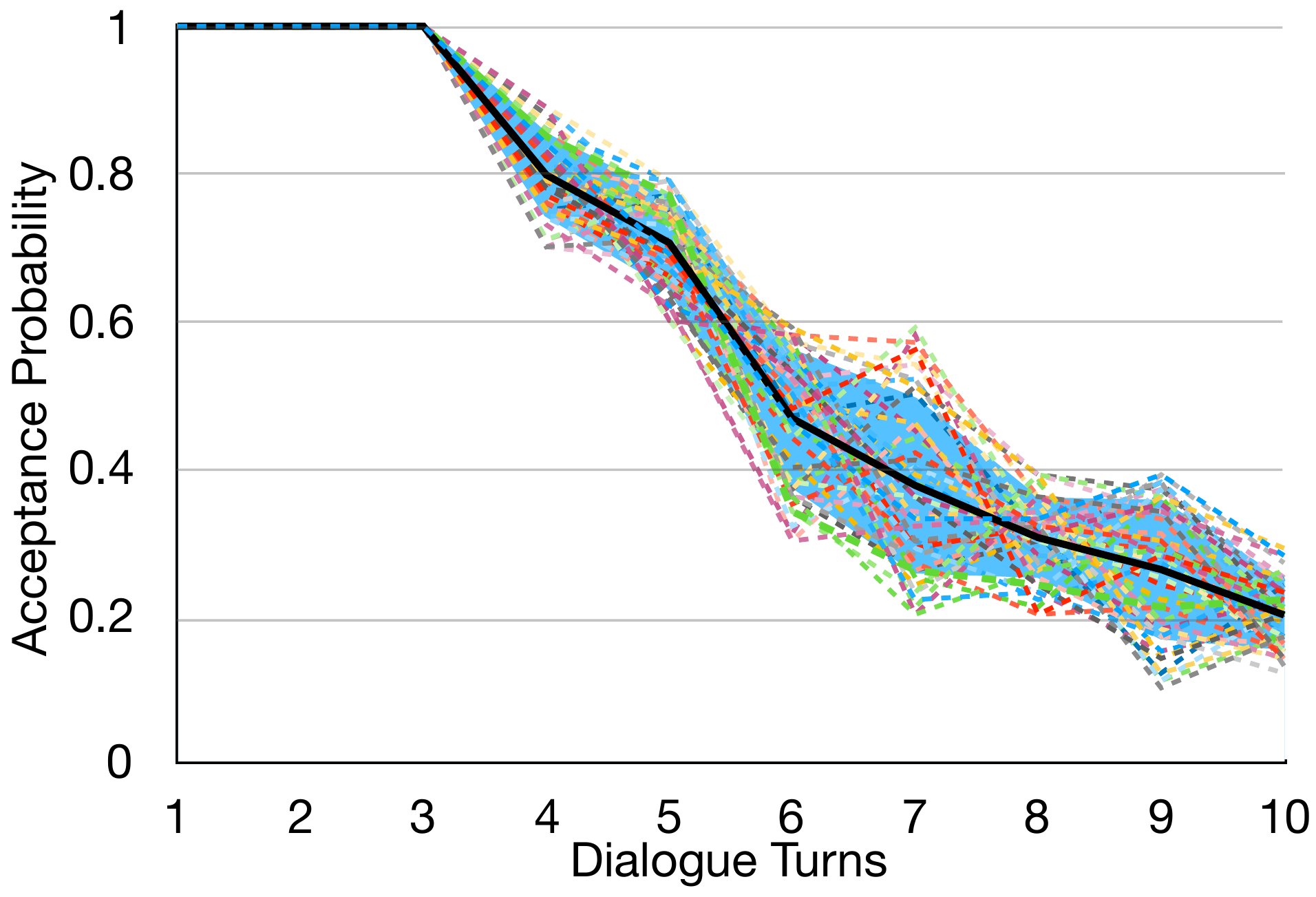}}
\caption{MCMC acceptance rate of the incremental parsing process. Dotted lines, black line, and blue shade are for samples, mean, and standard deviation, respectively.}
\label{fig:sample}
\end{center}
\vspace{-4mm}
\end{figure}

Figure \ref{fig:pg} also plots the average MCMC acceptance rate for the case, as defined in Formula \ref{formula_acc}, indicting the convergence of the inference. 
We see that the algorithm only needs to update the current graph belief slightly with a new perceived utterance. 
A peak in the curve can indicate that a key piece of information is detected that contradicts the existing belief: \textit{e.g.,} there is a peak of convergence curve in turn 7, which corresponds to \textit{``S5: Hey Mr. and Mrs. Geller!"}, indicating that S1 and S2 are a couple rather than friends. 
As such, we can see the algorithm get several relations updated accordingly.
We also show the convergence plots for 50 random testing cases from DialogRE in Figure \ref{fig:sample}, and the mean/standard deviation convergence rate as the black line/blue shade.
We prove that our updating algorithm is robust for the converged results. 
% Figure \ref{fig:pg} we show the parse graph sequence we inferred from an example dialogue. The inferred age of Emma actually helps the prediction of her relation with S3, so does the gender of S1. Also note that our method can do link prediction, which predicts the relation between two people who may not have direct conversation. For example, we infer S1 and S2 are spouse by their dialogues with other speakers. We also plot the average MCMC acceptance rate in Figure \ref{fig:sample} for the same example. We find that the acceptance rate decreases from a high level and the convergence is reached faster. Once we get a reasonable parse graph, we only need to update the parse graph slightly when a new utterance is perceived. A peak later in the dialogue usually indicates finding key information that contradicts the previous belief, \textit{e.g.,} in turn 6-7, utterance \textit{``Speaker 5: Hey Mr. and Mrs. Geller!"} indicates that S1 and S2 are spouse and several edges get updated accordingly.

\subsection{Ablation Study on $\alpha$--$\beta$--$\gamma$}
The $\alpha$--$\beta$--$\gamma$ strategy is designed to update relations and attributes jointly, having the input information flowing through the parse graph for the consistency of predictions. To validate the design, we ablate the processes on DialogRE to evaluate their impact on performance. Table \ref{tab:ablation} shows that $\alpha$ process, which is the discriminative model, makes the fundamental contribution, whereas $\beta$ and $\gamma$ processes alone cannot recognize social relations since they cannot perceive information from dialogues. Significantly, removing either one of the two processes will decrease the overall performance since the inference efficiency is reduced.
\begin{table}[]
\centering
\small
\begin{tabular}{lll|cc}
\toprule
\multicolumn{3}{c}{Processes} & \multirow{2}{*}{$\mathbf{F}1 (\sigma)$} & \multirow{2}{*}{$\mathbf{F}1_c (\sigma)$} \\
$\alpha$ & $\beta$  & $\gamma$ & & \\ 
\midrule
\checkmark &  & & 67.1 (0.5) & 64.2 (1.1)   \\
\checkmark &  \checkmark &  & 68.4 (0.8)  & 65.3 (0.6)   \\
\checkmark &  & \checkmark  & 68.3 (0.4)  & 65.2 (0.7)   \\
\checkmark  & \checkmark  & \checkmark    & \textbf{69.1 (0.5)}  & \textbf{66.5 (0.8)}   \\
\bottomrule
\end{tabular}
\caption{An ablation study on our parsing algorithm.}
\label{tab:ablation}
\vspace{-6mm}
\end{table}

\section{Conclusion}
The paper proposes a SocAoG model with $\alpha$--$\beta$--$\gamma$ processes for the consistent inference of social relations in dialogues. The model can also leverage attribute information to assist the inference. MCMC is proposed to parse the relation graph incrementally, enabling the dynamic inference upon any incoming utterance. Experiments show that our model outperforms state-of-the-art methods; case studies and ablation studies are provided for analysis. In the future, we will further explore how different initialization of the parse graph could help warm start the inference under various situations and how multi-modal cues could be leveraged.

\section*{Acknowledgments}
Y. W. is partially supported by NSF DMS 2015577. We would like to thank Yaofang Zhang and Qian Long for valuable discussions. We also thank the outstanding reviewers for their helpful comments to improve our manuscript. 

\section*{Ethical Considerations}
Endowing AI to understand social relations is an essential step towards building emotionally intelligent agents. By jointly inferring individual attributes and social relations, our incremental parsing algorithm enables consistent and dynamic relational inference in dialogue systems, which can be remarkably useful for a wide range of applications such as a chatbot that constantly perceives new information and conducts social relation inference. 

However, we never forget the other side of the coin. We emphasize that an ethical design principle must be in place throughout all stages of the development and evaluation. First, as discussed in \citet{larson-2017-gender}, we model the attributes as a social construct from a performative view. For example, ``gender performativity is not merely performance, but rather performances that correspond to, or are constrained by, norms or conventions and simultaneously reinforce them. Second, our model relies upon the attribute-category ascription provided by MovieGraph~\citep{vicol2018moviegraphs} and Friends Central in Fandom. However, we acknowledge that the annotation could be prone to a partial understanding of human relationships, and the real situation could be more complicated. Lastly, self-identification should be the gold standard for ascribing attribute categories. Practitioners are suggested to prompt users to provide self-identification and respect the difficulties of respondents when asking. Our model helps increase the interpretability of the relational inference process by tracking the attributes and updating the relational belief. We expect that the biases from relation recognition can be easier to measure, and our $\alpha$--$\beta$--$\gamma$ processes may provide a multidimensional way for correcting them.

\bibliographystyle{acl_natbib}
\bibliography{anthology,acl2021}

\appendix
\section{Appendices}
\label{sec:appendix}
% \begin{figure*}[ht]
% \begin{center}
% \centerline{\includegraphics[width=\linewidth]{all_type.pdf}}
% \caption{Performance boosts ($\mathbf{F}1$) of SocAoG compared to SimpleRE by all relation types.}
% \label{fig:full_type}
% \end{center}
% \vspace{-6mm}
% \end{figure*}

\begin{table*}[h]
\centering
\small
\resizebox{0.8\textwidth}{!}{%
\begin{tabular}{lllll}
\toprule
\textbf{ID} & \textbf{Subject} & \textbf{Relation Type}    & \textbf{Object}   & \textbf{Inverse Relation} \\
\midrule
1           & PER              & per:positive\_impression   & NAME              &                           \\
2           & PER              & per:negative\_impression   & NAME              &                           \\
3           & PER              & per:acquaintance          & NAME              & per:acquaintance          \\
4           & PER              & per:alumni                & NAME              & per:alumni                \\
5           & PER              & per:boss                  & NAME              & per:subordinate           \\
6           & PER              & per:subordinate           & NAME              & per:boss                  \\
7           & PER              & per:client                & NAME              &                           \\
8           & PER              & per:dates                 & NAME              & per:dates                 \\
9           & PER              & per:friends               & NAME              & per:friends               \\
10          & PER              & per:girl/boyfriend        & NAME              & per:girl/boyfriend        \\
11          & PER              & per:neighbor              & NAME              & per:neighbor              \\
12          & PER              & per:roommate              & NAME              & per:roommate              \\
13          & PER              & per:children              & NAME              & per:parents               \\
14          & PER              & per:other\_family          & NAME              & per:other\_family          \\
15          & PER              & per:parents               & NAME              & per:children              \\
16          & PER              & per:siblings              & NAME              & per:siblings              \\
17          & PER              & per:spouse                & NAME              & per:spouse                \\
18          & PER              & per:place\_of\_residence    & NAME              & gpe:residents\_of\_place    \\
19          & PER              & per:place\_of\_birth        & NAME              & gpe:births\_in\_place       \\
20          & PER              & per:visited\_place         & NAME              & gpe:visitors\_of\_place     \\
21          & PER              & per:origin                & NAME              &                           \\
22          & PER              & per:employee\_or\_member\_of & NAME              & org:employees\_or\_members  \\
23          & PER              & per:schools\_attended      & NAME              & org:students              \\
24          & PER              & per:works                 & NAME              &                           \\
25          & PER              & per:age                   & VALUE             &                           \\
26          & PER              & per:date\_of\_birth         & VALUE             &                           \\
27          & PER              & per:major                 & STRING            &                           \\
28          & PER              & per:place\_of\_work         & STRING            &                           \\
29          & PER              & per:title                 & STRING            &                           \\
30          & PER              & per:alternate\_names       & NAME/STRING       &                           \\
31          & PER              & per:pet                   & NAME/STRING       &                           \\
32          & GPE              & gpe:residents\_of\_place    & NAME              & per:place\_of\_residence    \\
33          & GPE              & gpe:births\_in\_place       & NAME              & per:place\_of\_birth        \\
34          & GPE              & gpe:visitors\_of\_place     & NAME              & per:visited\_place         \\
35          & ORG              & org:employees\_or\_members  & NAME              & per:employee\_or\_member\_of \\
36          & ORG              & org:students              & NAME              & per:schools\_attended      \\
37          & NAME             & unanswerable              & NAME/STRING/VALUE &              \\            
\bottomrule
\end{tabular}%
}
\caption{Relation types in DialogRE.}
\label{tab:dialogre_types}
\end{table*}

\begin{table*}[]
\centering
\resizebox{\textwidth}{!}{%
\begin{tabular}{|l|l|l|}
\hline
\multirow{4}{*}{attributes} & gender     & male, female                                                                                                                                                                                                                                                                                                                                                                                                                                                                                                                                                                                                                                                                                                                                                                                                                                                                                                                                                                                                                                                                                                                                                                                                                                                                                                                                                                                                                                                                                                                                                                                                                                                                                                                                                                                                                                                                                                                                                                                                                                                                                                                                                                                                                                                                                                                                                                                                                                                                                                                                                                                                                                                                                                                                                                                                                                                                                                                                                                                                                                                                                                                                                                                                                                                                                            \\ \cline{2-3} 
                            & age        & adult, kid, young adult, teenager, senior, baby                                                                                                                                                                                                                                                                                                                                                                                                                                                                                                                                                                                                                                                                                                                                                                                                                                                                                                                                                                                                                                                                                                                                                                                                                                                                                                                                                                                                                                                                                                                                                                                                                                                                                                                                                                                                                                                                                                                                                                                                                                                                                                                                                                                                                                                                                                                                                                                                                                                                                                                                                                                                                                                                                                                                                                                                                                                                                                                                                                                                                                                                                                                                                                                                                                                         \\ \cline{2-3} 
                            & ethnicity & caucasian, asian, arab, south-asian, hispanic, african, native american, other, aboriginal, african-american                                                                                                                                                                                                                                                                                                                                                                                                                                                                                                                                                                                                                                                                                                                                                                                                                                                                                                                                                                                                                                                                                                                                                                                                                                                                                                                                                                                                                                                                                                                                                                                                                                                                                                                                                                                                                                                                                                                                                                                                                                                                                                                                                                                                                                                                                                                                                                                                                                                                                                                                                                                                                                                                                                                                                                                                                                                                                                                                                                                                                                                                                                                                                                                            \\ \cline{2-3} 
                            & profession & \begin{tabular}[c]{@{}l@{}}photographer, cab driver, priest, writer, receptionist, delivery man, yoga instructor, chef, bartender, waitress, \\ tailor, parking attendant, student, professional, lawyer, teacher, businessman, secretary, model, prince, banker, \\ court reporter, intern, police officer, child psychologist, doctor, salesman/woman, hustler, bull rider, worker, \\ doctors, businessman/woman, nurse, barman, janitor, policeman, inspector, FDA agent, counselor, waiter, judge, \\ magician, prostitute, doorman, elevator operator, hotel manager, maid, bellhop, saleswoman, salesman, politician, \\ driver, usher, actress, actor, florist, pilot, flight attendant, film/tv producer, building manager, paramedic, federal agent, \\ postal worker, comic book artist, singer, executive, hockey player, referee, waiter/waitress, ex-soldier, receptionist, \\ mafia boss, mafia member, musician, drug lord, fruit vendor, barber, masseuse, mental patient, mental patient, \\ bus driver, night guard, housewife, editor, gardener, publisher, builder, elf, security guard, security chief, pedicurist, \\ professor of defense against the dark arts, wandmaker, wizard, caretaker, ghost, villain, Philadelphia Eagles fan, \\ cowboys America fan, bookmaker, unemployed, high school principal, jobless, racists, nuclear physicist, surgeon, \\ soldier, colonel, professor, engineer, military officer, technician, game show host, police, robber, waiter/waitress, \\ hitman, actor/actress, criminal, boxer, drug dealer, restaurant host, impersonator, military, trainer, manager, housekeeper, \\ veterinarian, sportsperson, sports coach, sports agent, accountant, personal assistant, nanny, reporter, tv host, cameraman, \\ tv presenter, cashier, artist, chauffeur, video artist, private investigator, administrator, tennis instructor, professional tennis player, \\ detective, ticket collector, director, medical workers, hospital orderly, pharmacist, security officer, dental assistant, dentist, \\ drug addict, registered sex offender, fetish worker, customer support, policemen, CEO, babysitter, assistant, principal, \\ guidance counselor, farmer, entertaining, domestic worker, fisherman, author, psychologist, security person, tv personality, \\ zeppelin crewman, king/queen, knight, journalist, assistant, weatherman, show host, make-up artist, seller, agent, tv show host, \\ makeup artist, treasure hunter, naval officer, steward, ship captain, ship designer, sailor, designer, carpenter, valet, bail bondsman, \\ court bailiff, court clerk, blackjack dealer, movie star, casino owner, casino manager, art director, executive recruiter, sports editor, \\ cowboy, cowboy employer, hacker, investment counselor, hairdresser, sports commentator, chemist, government rep, vicar, robot, \\ hotline agent, cook, surrogate date, philosopher, architect, record store owner, movie reviewer, call operator, bride, \\ dog sitter, newspaper employer, vet, insurance broker, union leader, tv reporter, senator, rancher, locksmith, district attorney, \\ store owner, smuggler, insurance agent, video editor, bouncer, trainee, real estate agent, prison guard, tour guide, mobster\end{tabular} \\ \hline
relations                   &            & \begin{tabular}[c]{@{}l@{}}sibling, parent, cousin, customer, friend, stranger, spouse, colleague, boss, would like to know, lover, mentor, engaged, \\ knows by reputation, acquaintance, roommate, best friend, antagonist, employed by, business partner, student, classmate, \\ patient, teacher, child, heard about, enemy, employer of, psychiatrist, doctor, collaborator, ex-lover, landlord, superior, \\ supervisor, grandchild, divorced, sponsor, ex-boyfriend, neighbor, fan, close friend, sister/brother-in-law, uncle, host, \\ employer, step-mother, foster-son, family friend, godfather, godson, brother-in-law, nanny, grandparent, aunt, aide, \\ students, family, customers, classmates, alleged lover, trainer, slave, hostage, robber, owner, instructor, competitor, \\ fiancee, aunt/uncle, mother-in-law, girlfriend, killer, babysitter, one-night stand, boyfriend, tenant, distant cousin, \\ father-in-law, mistress, agent, replacement, argue about the relationship, lawyer, ex-spouse, ex-girlfriend/ex-boyfriend, \\ niece/nephew, parent-in-law, guardian, operative system, couple, goddaughter, customer, ex-neighbor, worker, vet, \\ apprentice, public official, nurse, supporter, interviewee, interviewer, supporters, ex-fiance, fiance\end{tabular}                                                                                                                                                                                                                                                                                                                                                                                                                                                                                                                                                                                                                                                                                                                                                                                                                                                                                                                                                                                                                                                                                                                                                                                                                                                                                                                                                                                                                                                                                                                                                                                                                                                                                                                                                                                                                                                                                                                                                                                                     \\ \hline
\end{tabular}%
}
\caption{Attribute and relation types in MovieGraph.}
\label{tab:mg}
\end{table*}

\end{document}